\documentclass[10pt,twocolumn,letterpaper]{article}

\usepackage[final]{cvpr}
\usepackage[utf8]{inputenc}
\usepackage{times}
\usepackage{epsfig}
\usepackage{graphicx}
\usepackage{float}
\usepackage{amsmath}
\usepackage{breqn}
\usepackage{amssymb}
\usepackage{multirow}
\usepackage{booktabs}
\usepackage{hhline}
\usepackage{subcaption}
\usepackage[skip=1ex,font=small,labelsep=period]{caption}
\usepackage{enumitem}
\usepackage{tabularx}
\usepackage{mathtools}
\usepackage[dvipsnames]{xcolor}
\usepackage[pagebackref=true, colorlinks, linkcolor=red, citecolor=blue, urlcolor=blue]{hyperref}

\newcolumntype{Y}{>{\centering\arraybackslash}X}
\newcolumntype{R}{>{\raggedleft\arraybackslash}X}
\newcolumntype{L}{>{\raggedright\arraybackslash}X}

\makeatletter
\@namedef{ver@everyshi.sty}{}
\makeatother
\usepackage{tikz}

\usepackage[capitalize]{cleveref}
\crefname{section}{Sec.}{Secs.}
\Crefname{section}{Section}{Sections}
\Crefname{table}{Table}{Tables}
\crefname{table}{Tab.}{Tabs.}

\newcommand{\Modelname}{LISA}

\renewcommand{\vec}[1]{\boldsymbol{#1}}
\newcommand{\mat}[1]{\mathbf{#1}}

\newcommand{\real}[0]{\mathbb{R}}

\newcommand{\pose}[0]{\vec{\theta}}
\newcommand{\shape}[0]{\vec{\beta}}

\newcommand\customparagraph[1]{\vspace{0.7em}\noindent\textbf{#1}}

\newcommand\blfootnote[1]{%
  \begingroup
  \renewcommand\thefootnote{}\footnote{#1}%
  \addtocounter{footnote}{-1}%
  \endgroup
}

\newcommand{\bc}{\mathbf{c}}
\newcommand{\bd}{\mathbf{d}}

\newcommand{\bt}{\mathbf{t}}

\newcommand{\bv}{\mathbf{v}}

\newcommand{\bbv}{\bar{\mathbf{v}}}
\newcommand{\bw}{\mathbf{w}}
\newcommand{\bx}{\mathbf{x}}

\newcommand{\bJ}{\mathbf{J}}

\newcommand{\bR}{\mathbf{R}}

\newcommand{\bT}{\mathbf{T}}

\newcommand{\bV}{\mathbf{V}}

\newcommand{\bW}{\mathbf{W}}

\newcommand{\bhc}{\hat{\mathbf{c}}}

\newcommand{\bhJ}{\hat{\mathbf{J}}}

\newcommand{\bhw}{\hat{\mathbf{w}}}

\newcommand{\bgamma}{\boldsymbol{\gamma}}

\def\cvprPaperID{6216} %
\def\confName{CVPR}
\def\confYear{2022}

\begin{document}

\title{LISA: Learning Implicit Shape and Appearance of Hands}

\author{
Enric Corona$^{1\dagger}$
\hspace{2ex} Tomas Hodan$^{2}$
\hspace{2ex} Minh Vo$^{2}$
\hspace{2ex} Francesc Moreno-Noguer$^{1}$ \\
\hspace{2ex} Chris Sweeney$^{2}$ 
\hspace{2ex} Richard Newcombe$^{2}$
\hspace{2ex} Lingni Ma$^{2}$\\ \vspace{1ex}
\normalsize{${}^{1}$Institut de Robòtica i Informàtica Industrial, CSIC-UPC, Barcelona, Spain} \hskip3ex \normalsize{${}^{2}$Reality Labs, Meta}
}

\input{figure_teaser}

\blfootnote{$\dagger$ Work performed during internship with Reality Labs, Meta.}

\begin{abstract}
This paper proposes a do-it-all neural model of human hands, named LISA. The model can capture accurate hand shape and appearance, generalize to arbitrary hand subjects, provide dense surface correspondences, be reconstructed from images in the wild, and can be easily animated.
We train LISA by minimizing the shape and appearance losses on a large set of multi-view RGB image sequences annotated with coarse 3D poses of the hand skeleton.
For a 3D point in the local hand coordinates, our model predicts the color and the signed distance with respect to each hand bone independently, and then combines the per-bone predictions using the predicted skinning weights. The shape, color, and pose representations are disentangled by design, enabling fine control of the selected hand parameters.
We experimentally demonstrate that LISA can accurately reconstruct
a dynamic hand from monocular or multi-view sequences, achieving a noticeably higher quality of reconstructed hand shapes compared to baseline approaches.
Project page: \url{https://www.iri.upc.edu/people/ecorona/lisa/}.

\end{abstract}

\section{Introduction}

Since the thumb opposition enabled grasping around 2 million years ago~\cite{karakostis2021biomechanics}, humans interact with the physical world mainly with hands. The problems of modeling and tracking human hands have therefore naturally received a considerable attention in computer vision~\cite{oikonomidis2018hands18}. Accurate and robust solutions to these problems would unlock a wide range of applications in, \eg, human-robot interaction, prosthetic design, or virtual and augmented reality.

Most research efforts related to modeling and tracking human hands, \eg,~\cite{freihand:iccv19, obman2:hasson:cvpr20, interhand:moon:eccv20, ho3d:hampali:cvpr20, dexycb:chao:cvpr21, liu:cvpr21}, rely on the MANO hand model~\cite{mano:romero:tog17}, which is defined by a polygon mesh that can be controlled by a set of shape and pose parameters. Despite being widely used, the MANO model has a low resolution and does not come with texture coordinates, which makes representing the surface color difficult.

The related field of modeling and tracking human bodies has been relying on parametric meshes as well, with the most popular model being SMPL~\cite{smpl:loper:tog15} which suffers from similar limitations as the MANO model. Recent approaches for modeling human bodies, \eg,~\cite{loopreg:bhatnagar:2020,nasa:deng:2020,ptf:shaofei:2021,imghum:allidieck:iccv21,snarf:chen:iccv21,leap:mihajlovic:cvpr21,scanimate:saito:cvpr21}, rely on articulated models based on implicit representations, such as Neural Radiance Field~\cite{nerf:eccv20} or Signed Distance Field (SDF)~\cite{deepsdf:park:cvpr19}. Such representations are capable of representing both  shape and  appearance and able to capture finer geometry compared to approaches based on parametric meshes. However, it is yet to be explored how well  implicit representations apply to articulated objects such as the human hand and how they generalize to unseen poses.

We explore articulated implicit representations for modeling human hands and make the following contributions:

\begin{enumerate}[topsep=0pt,itemsep=-1ex,partopsep=1ex,parsep=1ex]
    \item We introduce LISA, the first neural
    model of human hands that can capture accurate hand shape and appearance, generalize to arbitrary hand subjects, provide dense surface correspondences (via predicted skinning weights), be reconstructed from images in the wild, and easily animated.
    \item We show how to train LISA by minimizing shape and appearance losses on a large set of multi-view RGB image sequences annotated with coarse 3D poses of the hand skeleton.
    \item The shape, color and pose representations in LISA are disentangled by design, enabling fine control of selected aspects of the model.
    \item Our experimental evaluation shows that LISA surpasses baselines in hand reconstruction from 3D point clouds and hand reconstruction from RGB images.
    
\end{enumerate}

\begin{figure*}[t!]
\noindent\includegraphics[width=1.0\linewidth]{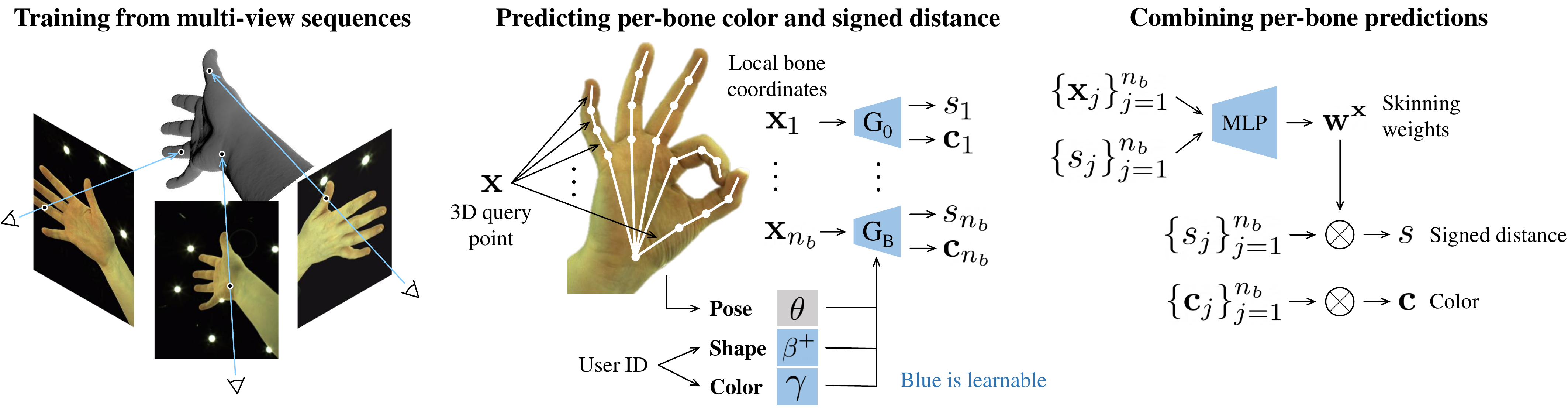}
\caption{
\textbf{Training and architecture of the LISA hand model.} \emph{Left:} LISA is trained by minimizing shape and appearance losses from a dataset of multi-view RGB image sequences. The sequences are assumed annotated with coarse 3D poses of the hand skeleton that are refined during training.
The training sequences show hands of multiple people and are used to learn disentangled representations of pose, shape and color.
\emph{Middle:} LISA approximates the hand by a collection of rigid parts defined by the hand bones. A 3D query point is transformed to the local coordinate systems of the bones associated with independent neural networks, which predict the signed distance to the hand surface and the color. Note that $G_j$ is realized by two independent MLPs, one predicting the signed distance and one predicting the color (see Section~\ref{sec:model_definition}). \emph{Right:} The per-bone predictions are combined using skinning weights predicted by an additional network.
\label{fig:diagram}
}
\end{figure*}

\section{Related work}

\noindent\textbf{Parametric meshes.}
Thanks to their simplicity and efficiency, parametric meshes gained great popularity
for modeling articulated objects such as bodies~\cite{smpl:loper:tog15, smplx:pavlakos:2019, frank:joo:cvpr18, star:osman:eccv20}, hands~\cite{mano:romero:tog17}, faces~\cite{flame:li:2017} and animals~\cite{smal:zuffi:cvpr17}.
The MANO hand model~\cite{mano:romero:tog17} is learned from a large set of carefully registered hand scans and captures shape-dependent and pose-dependent blend shapes for hand personalization.
Despite widely adopted in hand tracking and shape estimation~\cite{obman:hasson:cvpr19, hamr:zhang:iccv19, obman2:hasson:cvpr20, freihand:iccv19, interhand:moon:eccv20, ho3d:hampali:cvpr20, i2lmesh:moon:eccv20, dexycb:chao:cvpr21, liu:cvpr21, h2o:kwon:iccv21, rho:cao:iccv21, ganhand}, the MANO mesh is limited by a low resolution rooted from solving a large optimization problem with classical techniques.
To reconstruct finer hand geometry, graph convolutional networks are explored in~\cite{gcnn:ge:cvpr19, pose2mesh:choi:eccv20, gcnn:tang:iccv21} and spiral filters in~\cite{spiral:kulon:cvpr20}. 
Based on a professionally designed mesh template, DeepHandMesh~\cite{dhm:moon:eccv20} learns the pose and shape corrective parameters by a neural network.
Chen~et.~al.,~\cite{i2uv:chen:iccv21} refined MANO by developing a UV-based representation.
GHUM~\cite{ghum:xu:cvpr20} introduces a generative parametric mesh where the shape corrective parameters, skeleton and blend skinning weights are predicted by a neural network.

\customparagraph{Implicit shape representations.}
Many works adopt neural networks to model geometry by learning an implicit function, which is continuous and differentiable, such as the signed distance field (SDF)~\cite{deepsdf:park:cvpr19, lif:chen:cvpr19, igr:gropp:icml20, sal:atzmon:cvpr20, sald:atzmon:iclr21, corona2021smplicit} or the occupancy field~\cite{occnet:mescheder:cvpr19}. 
To improve learning efficiency, \cite{sif:genova:2019, ldif:genova:2020, patchnet:eccv20, dls:chabra:eccv20} studied part-based implicit templates to model mid-level object-agnostic shape features.
Implicit representations were extended to articulated deformation,
in LoopReg~\cite{loopreg:bhatnagar:2020} with a weakly-supervised training using cycle consistency by learning inverse skinning, which maps surface points to the SMPL human body model~\cite{smpl:loper:tog15}.
Based on SMPL,
NASA~\cite{nasa:deng:2020} trains one OccNet~\cite{occnet:mescheder:cvpr19} per skeleton bone to approximate the shape blend shapes and pose blend shapes.
PTF~\cite{ptf:shaofei:2021} extends NASA and registers point clouds to SMPL.
In a similar spirit, imGHUM~\cite{imghum:allidieck:iccv21} trains four DeepSDF networks~\cite{deepsdf:park:cvpr19} whose predictions are fused by an additional lightweight network. %
To eliminate the need of having the ground-truth SMPL in NASA training, SNARF~\cite{snarf:chen:iccv21} utilizes an iterative root finding technique to
link every query point in the posed space to the corresponding point in the canonical space, which enables differentiable forward skinning. 
LEAP~\cite{leap:mihajlovic:cvpr21} and SCANimate~\cite{scanimate:saito:cvpr21} additionally model both forward and inverse skinning by a neural network, and use cycle consistency to supervise training of transformation to the canonical space. LEAP also extends the framework to multi-subject learning by mapping the bone transformation to a shape feature, and SCANimate builds animatable customized clothed avatars.
We take inspiration from NASA to constrain hand deformation, but explicitly model the skinning weights for blending shape and color.

\customparagraph{Implicit appearance representations.}
A number of approaches have been proposed to learn
appearance of a scene from multi-view images. 
The idea is to model the image formation process by rendering a neural volume with ray-casting~\cite{srn:sitzmann:nips19, nv:lombardi:tog19, dvr:cvpr20, idr:yariv:nips20}.
Particularly, NeRF~\cite{nerf:eccv20} gains popularity with an efficient formulation of modeling the radiance field. 
Follow-up studies show that geometry can be improved if the density is regulated by occupancy~\cite{unisurf:oeschsle:iccv21} or SDF~\cite{neus:wang:nips21, volsdf:yariv:arxiv21}. In this work, we use VolSDF~\cite{volsdf:yariv:arxiv21} as a backbone renderer.
For dynamic scenes, \cite{dnerf:cvpr20, nerfies:iccv21,nrnerf:iccv21} combine NeRF with learning a deformation field. For modeling dynamic human bodies,
Neural Body~\cite{neuralbody:cvpr21} attaches learnable vertex features to SMPL, and diffuses the features with sparse convolution for volumetric rendering. A-NeRF~\cite{anerf:nips21} conditions NeRF with SMPL bone transformations to learn an animatable avatar. Similar ideas are proposed in~\cite{animatable:iccv21} and NARF~\cite{narf:iccv21}.
H-NeRF~\cite{hnerf:nips21} combines imGHUM with NeRF to enable appearance learning and train a separate network to predict SDF. In our work, the prediction of appearance and SDF is independent within each bone and later weighted by the corresponding skinning weights.

\customparagraph{Disentangled representations.} 
Disentangling parameters of certain properties such as pose, shape or color is desireable as it allows treating (\eg, estimating or animating) these properties independently.
Inspired by the parametric mesh models, Zhou et al.,~\cite{usp:zhou:eccv20} trained a mesh auto-encoder to disentangle shape and pose of humans and animals. They developed an unsupervised learning technique based on a cross-consistency loss.
DiForm~\cite{diform:xu:arxiv21} adopted a decoder network to disentangle identity and deformation in learning an SDF-based shape embedding.
A-SDF~\cite{asdf:mu:iccv21} factored out shape embedding and joint angles to model articulated objects.
NPM~\cite{npm:palafox:iccv21} proposed to train shape embedding on canonically posed scans, followed by another network to learn the deformation field with dense supervision. 
A similar idea to the deformation field was adopted by i3DMM~\cite{i3dmm:cvpr21} to learn a human head model. The method disentangles identity, hairstyle, and expression and is trained with dense colored SDF supervision.
In this work, we propose a generative hand representation with disentangled shape, pose and appearance parameters.

\section{Background}\label{sec:background}

\noindent\textbf{MANO}~\cite{mano:romero:tog17} represents the human hand as a function of the pose parameters $\pose$ and shape parameters $\shape$:
\begin{equation}
    M: (\pose,\shape) \mapsto \bV\label{eq:mano}\;,
\end{equation}
where the hand is defined by a skeleton rig with $n_j=16$ joints, and the pose parameters $\pose \in\real^{n_j\times 3} $ represent the axis-angle representation of the relative rotation between bones of the skeleton. $\shape$ is a 10-dimensional vector and $\bV\in\real^{n_v\times 3}$ are the vertices of a triangular mesh. The mapping $M$ is estimated by deforming a canonical hand $\bV^r$ by a Linear Blend Skinning (LBS) transformation, with weights $\bW\in\real^{n_b\times n_v}$, where $n_b$ is the number of bones. Concretely, given a vertex $\bv^r_i$ on the canonical shape, LBS transforms the vertex as follows:
\begin{equation}
    \bv_i = \sum_{j=1}^{n_b} w_{i,j}\bT_j\bbv^r_i\;,
\end{equation}
where $\bT_j\in\real^{3\times 4}$ is the rigid transformation applied on the rest pose of bone $j$, $w_{i,j}$ is the $(i,j)$ entry of $\bW$ and $\bbv$ denotes the homogeneous coordinates of $\bv$.
LISA builds on MANO's definition of the skeleton by using the same pose parameters $\pose$ and bone transformations.

\customparagraph{NeRF/VolSDF}. 
NeRF~\cite{nerf:eccv20} is a state-of-the-art rendering algorithm for novel view synthesis. The algorithm models the continuous radiance field of a static scene by learning the following function:
\begin{equation}
F:(\bx,\bd) \mapsto (\bc,\sigma)\;,  
\end{equation}
which maps a 3D location $\bx\in\mathbb{R}^3$ and the viewing direction $\bd\in\mathbb{R}^2$ passing through $\bx$ to the color value $\bc\in\mathbb{R}^3$ and its density $\sigma\in\mathbb{R}$. The function $F$ is modeled by an Multilayer Perceptron (MLP) network, which is trained from a set of dense multi-view posed RGB images of a single static scene. 
While NeRF has shown impressive novel view synthesis results, the estimated volume density is however not effective to infer accurate geometry. A number of recent works studied this problem~\cite{unisurf:oeschsle:iccv21, volsdf:yariv:arxiv21, neus:wang:nips21} and propose to extended NeRF by incorporating SDF~\cite{deepsdf:park:cvpr19}. In this paper we adapt the formulation of VolSDF~\cite{volsdf:yariv:arxiv21}, which defines the volume density as a Laplace's Cumulative Distribution Function (CDF) applied to a SDF representation. VolSDF also disentangles the geometry and appearance learning using two MLPs for SDF and color estimation, respectively. %

\section{LISA: The proposed hand model}\label{sec:approach}

This section provides a detailed description of the proposed hand model, which we dub \Modelname~for Learning Implicit Shape and Appearance model.

\customparagraph{Problem settings.}
Consider a dataset of multi-view RGB video sequences with known camera calibration. Each sequence captures a single hand from a random person posing random motion. The objective is to learn a hand model, which reconstructs the hand geometry, the deformation and the appearance, while also generalizes to reconstruct unseen hands and motion from test images. In contrast to prior hand modeling works, which often require a large collection of high-quality 3D hand scans, we consider a setup that lowers the requirement for data collection but adds the challenge to the algorithm.
Inspired by classical hand modeling approaches, we assume that a kinematic skeleton is associated with the hand, where the coarse 3D poses are produced by pre-processing the training sequences with the state-of-art hand tracking. The motivation of using a skeleton is to regulate the hand deformation with articulation and to enable animation for the obtained model. To focus the deep network on the hands, we further simplify the input by assuming the foreground masks are known.

\subsection{Model definition}\label{sec:model_definition}
Our goal is to learn a mapping function from the parametric skeleton to a full hand model of the shape and appearance. In this work, we choose the skeleton to be parameterized by MANO and formulate the learning as:
\begin{equation}
    M^+: (\pose,\shape^+,\bgamma) \mapsto \boldsymbol{\psi}\;,
    \label{eq:model}
\end{equation}
which maps the pose parameter $\pose$, shape parameter $\shape^+$ and the color parameter $\bgamma$ to an implicit representation $\boldsymbol{\psi}$. Here, we indicate the shape parameter is different from that of MANO using the superscript $^+$. The implicit representation $\boldsymbol{\psi}$ is a continuous function for the geometry and appearance. Similar to radiance field definition, this is defined as: 
\begin{equation}\label{eq:implicit}
    \boldsymbol{\psi}: (\bx, \mathbf{d}) \mapsto (s, \bc) \;,
\end{equation}
which returns the SDF value $s\in\mathbb{R}$ and the color value $\bc$ for a query 3D point $\bx$ and the view direction $\mathbf{d}$. Using the implicit representation, the learned model is not tied to template meshes with fixed resolution, and therefore can encode detailed deformation more efficiently. The hand surface is represented by the $0$-level set of~$s$, where the 3D mesh $\bV$ can be extracted by uniformly sampling the 3D space and applying Marching Cubes~\cite{marchingcubes}. Putting \cref{eq:model} and \cref{eq:implicit} together, and with removing the viewing direction for simplified notation, yield the mapping we aim to learn:
\begin{equation}\label{eq:mapping}
    G: (\bx,\pose,\shape^+,\bgamma) \mapsto (s,\bc)\;.
\end{equation}
In the remainder of the section, we explain how to model \cref{eq:mapping} with network training.

\customparagraph{Independent per-bone predictions with skinning.}
Following
\cite{nasa:deng:2020, neuralparts}, we approximate the overall hand shape by a collection of rigid parts, which are in our case defined by $n_b$ bones.
Specifically, the network $G$ is split into $n_b$ MLPs predicting the signed distance, $\{G^s_j\}_{j=1}^{n_b}$, and $n_b$ MLPs predicting the color, $\{G^c_j\}_{j=1}^{n_b}$, with each MLP making an independent prediction with respect to one bone.
As the input images correspond to posed hands, the point $\bx$ is first unposed (\ie, transformed to the coordinate space of the hand in the rest pose)
using the kinematic transformations of bones, $\{\bT_j\}_{j=1}^{n_b}$:
$\bx_j=\bR_j^{-1}(\bx - \bt_j )$, where $\bR_j$ and $\bt_j$ are the rotation and translation components of the transformation $\bT_j$.
With this formulation, we collect a set of independent SDF predictions and color predictions for each query point $\{s_j,\bc_j\}_{j=1}^{n_b}$, where:
\begin{align}
    & G^s_j: (\bx_j,\pose,\shape^+,\bgamma) \mapsto s_j\;, \\
    & G^c_j: (\bx_j,\pose,\shape^+,\bgamma) \mapsto \bc_j\;.
\end{align}
To combine the per-bone output into a single SDF and color addition, we introduce an additional MLP to learn the weights. The weight MLP takes the input as the concatenation of $n_b$ unposed $\bx_j$ and the predicted SDF $s_j$ per MLP 
, to output the weighting vector $\bw=[w_1,\ldots,w_{n_b}]$. A softmax layer is used to constrain the value of $\bw$ to be probability-alike, i.e., $w_i \geq 0\;,\forall i$ and $\sum_i w_i = 1$. 
The final output for a query point is then computed by:
\begin{equation}
    s = \sum_{j=1}^{n_b} w_j s_j\;,\hspace{2mm}\bc = \sum_{j=1}^{n_b} w_j \bc_j\;.
\end{equation}

Note the weight vector $\bw$ is an analogy to skinning weights in classic LBS-based models. The similar design has also been explored by NASA~\cite{nasa:deng:2020} and NARF~\cite{narf:iccv21}. The difference is that NASA selects one MLP output, which is determined by the maximum of the predicted occupancies. NARF proposes to learn the weights with an MLP, but only uses the canonicalized points to train this module. In our design, the network sees both canonicalized points and the per-bone SDF. The SDF serves as a valuable guide in learning skinning weight. More importantly, the gradients can now back-propagate via the weights to train the per-bone MLPs. This means MLPs can leverage $\bw$ to avoid learning SDF for far-away points. We show in experiments that this design greatly improves geometry.

\customparagraph{Model rendering.}
As in~\cite{volsdf:yariv:arxiv21}, we first need to obtain the volume densities from the predicted signed distance field before rendering. We infer it indirectly from the predicted signed distances:
\begin{equation}
    \sigma(\bx)=\alpha\Psi_\beta(-s)\;,
\end{equation}
where $s$ is the signed distance of $\bx$, $\Psi_\beta(\cdot)$ is the CDF of the Laplace distribution, and $\alpha$ and $\beta$ are two learnable parameters (see~\cite{volsdf:yariv:arxiv21} for further details).

The color of a specific image pixel is then estimated via the volume rendering integral, by accumulating colors and volume densities along its corresponding camera ray $\mat{d}$. In particular, the color of the pixel $\hat{\mat{c}}_k$ is approximated by a discrete integration between near and far bounds $t_n$ and $t_f$ of a camera ray $\mat{r}(t) = \mat{o} + t\mat{d} $ with origin $\mat{o}$:
\begin{equation}
    \bc_k = \int_{t_n}^{t_f} T(t) \sigma(r(t))\mat{c}(\mat{r}(t), ) dt\,,
\end{equation}
where:
\begin{equation}
    T(t) = \exp \left( -\int_{t_n}^{t}\sigma(\mathbf{r}(s))ds \right)\,.
\end{equation}

\subsection{Training} \label{subsec:training}
As shown in Fig.~\ref{fig:diagram}, the parameters of \Modelname~that need to be learned are: (1) the MLPs for predicting signed distance and color for the $n_b$ bones, (2) the MLP that estimates the skinning weights, and (3) the shape $\shape^+$ and color $\bgamma$ latent codes to control the generation process. Note that the pose $\pose$ is not learned and assumed given during training.
We next explain how we learn these parameters from the multi-view image sequences from the InterHand2.6M dataset~\cite{interhand:moon:eccv20}. 

\customparagraph{Disentangling shape, color and pose.} \Modelname~is designed to completely disentangle the representations of pose, shape and color.
The shape $\shape^+$ and color $\bgamma$ parameters are fully learnable latent vectors. Since both are user specific, we assign the same latent code for all images of the same person. In both cases, they are represented as 128-dimensional vectors, initialized from a zero-mean multivariate-Gaussian distribution with a spherical covariance, and optimized during training following the auto-decoder formulation of~\cite{deepsdf:park:cvpr19}.

The pose parameters $\pose$ are defined by the 48-dimensional representation of MANO. When training on InterHand2.6M, we kept the provided ground-truth pose parameters fixed for the initial $10\%$ of training steps, then we started optimizing the parameters to account for errors in the ground-truth annotations.

\customparagraph{Color calibration.} 
In order to allow for slight differences in the intensity of the training images, we follow Neural Volumes~\cite{nv:lombardi:tog19} and introduce a per-camera and per-channel gain $g$ and bias $b$ that is applied to the rendered images at training time. At inference, we use the average of these calibration parameters.

\customparagraph{Loss functions.}
To learn \Modelname, we minimize a combination of losses that aim to ensure accurate representation of the hand color while properly regularizing the learned geometry. Specifically, we optimize the network by randomly sampling a batch of viewing directions $\bd_k$ and estimating the corresponding pixel color via volume rendering. Let $\bc_k$ be the estimated pixel color and $\bhc_k$ the ground truth value. The first loss we consider is:
\begin{equation}
    \mathcal{L}_{\textrm{col}}=\|\bc_k - \bhc_k \|_1\;,
    \label{eq:color_loss}
\end{equation}
where $\|\cdot\|_j$ denotes the $j$-norm.
We also regularize the SDF of $G(\cdot)$ with the Eikonal loss~\cite{igr:gropp:icml20} to ensure it approximates a signed distance function:
\begin{equation}
    \mathcal{L}_{\textrm{Eik}} = \sum_{\bx \in \Omega }(\|\nabla_{\bx}G(\bx)\|_2 -1 )^2\;,
    \label{eq:eik_loss}
\end{equation}
where $\Omega$ is a set of points sampled both on the surface and uniformly taken from the entire scene.
In order to prevent local minima in regions relying only on one or a few bones,
We use the pseudo-ground truth pose and shape parameters to obtain an approximate 3D mesh and its corresponding skinning weights $\bhw$, which we use to supervise the predicted skinning weights~$\bw$:
\begin{equation}
    \mathcal{L}_{\textrm{\bw}}=\|\bw - \bhw \|_1\;.
\end{equation}

\noindent Finally, we also regularize the latent vectors $\shape^+$  and $\bgamma$:
\begin{equation}
    \mathcal{L}_{\textrm{reg}}=\|\shape^+\|_2 + \|\bgamma\|_2. 
    \label{eq:regulariation}
\end{equation}

\noindent The full loss
is a linear combination of the four previous loss terms (with hyperparameters $\lambda_{\textrm{col}}$, $\lambda_{\textrm{Eik}}$, $\lambda_{\textrm{\bw}}$ and $\lambda_{\textrm{reg}}$):
\begin{equation}
    \mathcal{L}= \lambda_{\textrm{col}}\mathcal{L}_{\textrm{col}} + \lambda_{\textrm{Eik}}\mathcal{L}_{\textrm{Eik}} + \lambda_{\textrm{\bw}}\mathcal{L}_{\textrm{\bw}} + \lambda_{\textrm{reg}}\mathcal{L}_{\textrm{reg}}\;.
    \label{eq:loss_total}
\end{equation}

\customparagraph{Learning a prior for human hand SDFs.} 
When minimizing \cref{eq:loss_total}, we face two main challenges. First, since we only supervise on images, the simultaneous optimization of shape and texture parameters may lead to local minima with good renders but
wrong geometries. Second, the InterHand2.6M dataset~\cite{interhand:moon:eccv20} we use for training has a large number of images ($\sim$130k) but they only correspond to 27 different users, compromising the generalization of the model.

To alleviate these problems, we build a shape prior using the 3DH dataset~\cite{diform:xu:arxiv21}, which contains $\sim$13k 3D posed hand scans of 183 different users. The scans are used to pre-train the geometry MLPs in $G(\cdot)$, which we denote $G_{\beta^+}(\cdot)$, and which are responsible for predicting the signed distance~$s$:
\begin{equation}
    G_{\beta^+}: (\bx,\pose,\shape^+) \mapsto s\;.
\end{equation}
We pre-train $G_{\beta^+}$ with two additional losses. First, assuming $\bx_{\textrm{surf}}$ to be a point of a 3D scan, we enforce $G_{\beta^+}$ to predict a $0$ distance on that point: 
\begin{equation}
    \mathcal{L}_{\textrm{surf}}=\|G_{\beta^+}(\bx_{\textrm{surf}},\pose,\shape^+)      \|_1.
\end{equation}
We also supervise the gradient of the signed distance with the ground truth normal  $N(\bx_{\textrm{surf}})$ at $\bx_{\textrm{surf}}$:
\begin{equation}
    \mathcal{L}_{\textrm{N}} =  \|\nabla_{\bx_{\textrm{surf}}}G(\bx_{\textrm{surf}}) -N(\bx_{\textrm{surf}}) \|_1\;,
\end{equation}
where $N(\bx)$ is the  3D normal direction at $\bx$.

With these two losses, jointly with losses $\mathcal{L}_{\textrm{Eik}}$, $\mathcal{L}_{\textrm{\bw}}$ and the regularization  $\|\shape^+\|_2$, we   learn a prior on $\shape^+$ which is used to initialize the full optimization of the model in \cref{eq:loss_total}. As we show in the experimental section, this prior allows to significantly boost the performance of \Modelname.

\subsection{Inference}
In the experimental section we apply the learned model to 3D reconstruction from point-clouds and to 3D reconstruction from images. Both of these applications involve an optimization scheme which we describe below.

\customparagraph{Reconstruction from point clouds.} 
Let $\mathcal{P}=\{\bx_i\}_{i=1}^{n}$ be a point-cloud with $n$ 3D points.  To fit our trained model to this data, we follow a very similar pipeline as the one used to learn the prior. Specifically, we minimize the following objective function:
\begin{equation}
    \mathcal{L}(\pose,\shape^+)= \sum_{\bx \in \mathcal{P}}\|G_{\shape^+}(\bx,\pose,\shape^+)\|_1 + \|\shape^+\|_2\;.
\end{equation}

\customparagraph{Reconstruction from monocular or multiview images.} 
Given an input image $\mathcal{I}$, we assume we have a coarse foreground mask and that the 2D locations of $n_j$ hand joints, denoted as $\bhJ^{\textrm{2D}}$, are available. These locations can be detected using, \eg, OpenPose~\cite{openpose}. To fit \Modelname~to this data, we minimize the following objective:
\begin{equation}
    \mathcal{L}(\pose,\shape^+)=
    \sum_{\bd\in \mathcal{I}}\mathcal{L}_{\textrm{col}}(\bd) +   \mathcal{L}_{\textrm{reg}} +
    \mathcal{L}_{\textrm{joints}}\;,
\end{equation}
where the first two terms correspond to the color loss of \cref{eq:color_loss}  (expanded to all viewing directions intersecting the pixels of the input image), and the shape and pose regularization loss in \cref{eq:regulariation}. The last term is a joint-based data term that penalizes the 2D distance between the estimated 2D joints and the projected  3D joints $\bJ^{\textrm{3D}}$ computed from the estimated pose parameters $\pose$: 
\begin{equation}
    \mathcal{L}_{\textrm{joints}}=\|\bhJ^{\textrm{2D}} - \pi(\bJ^{\textrm{3D}})\|_1\;,
\end{equation}
where $\pi(\cdot)$ is the 3D-to-2D projection. We also use extrinsic camera parameters in case of multi-view reconstruction.

\section{Experiments}\label{sec:exp}

In this section, we evaluate \Modelname~on the tasks of hand reconstruction from point clouds and hand reconstruction from RGB images, and
demonstrate that
it outperforms the state of the art by a considerable margin.

\setlength{\tabcolsep}{0pt}
\begin{table}[t!]
\footnotesize
\begin{center}
\begin{tabularx}{\columnwidth}{l Y Y Y Y}
\toprule
& \multicolumn{2}{c}{Reconstruction to scan} & \multicolumn{2}{c}{Scan to reconstruction} \\

\cmidrule{1-5}

Method &
V2V [mm] &
V2S [mm] &
V2V [mm] &
V2S [mm] \\

\cmidrule{1-5}
\multicolumn{5}{c}{3DH dataset~\cite{diform:xu:arxiv21}:} \\
\cmidrule{1-5}

MANO~\cite{mano:romero:tog17} & 3.27 & 2.11 & 3.44 & 3.23 \\
VolSDF~\cite{deepsdf:park:cvpr19} & 3.69 &	1.26  & 5.33 & 	5.23	 \\
NASA~\cite{nasa:deng:2020} & 3.05 & 	1.14 &3.69 & 	3.66	  \\
NARF~\cite{narf:iccv21} & 4.69 & 	2.19 & 2.05 & 	2.01	 \\
LISA-im & {\bf 2.93} & 	{\bf 0.93} & {\bf 1.90 } & 	{\bf 1.87}	 \\
\cmidrule{1-5}
LISA-geom & 0.83 &  0.43 &  0.63 &  0.54  \\
LISA-full & 1.93 & 0.63 & 1.50 & 1.43  \\

\cmidrule{1-5}
\multicolumn{5}{c}{MANO dataset~\cite{mano:romero:tog17}:} \\
\cmidrule{1-5}

MANO~\cite{mano:romero:tog17} & 3.14 & 2.92 & 3.90 & 1.57  \\
VolSDF~\cite{deepsdf:park:cvpr19} & 3.69 &	2.22  & 2.37 &	2.23 \\
NASA~\cite{nasa:deng:2020} & 	5.31 &	3.80  & 2.57 &	2.33 \\
NARF~\cite{narf:iccv21} & 4.02 &	2.69  & 2.11 &	2.06	  \\
LISA-im & {\bf 3.09} &	{\bf 1.96 } & {\bf 1.19} & {\bf 1.13}	  \\
\cmidrule{1-5}
LISA-geom &  0.36 & 0.16  & 0.81 & 0.26  \\
LISA-full &  1.45 & 0.64  & 0.64 & 0.58  \\

\bottomrule
\end{tabularx}
\caption{\small{\textbf{Shape reconstruction from point clouds.}
The 3D shape reconstructions are evaluated by the vertex-to-vertex and vertex-to-surface distances (in mm). LISA-im is consistently superior among methods trained on images only. Using the geometric prior (LISA-geom, LISA-full) yields a significant boost in performance. %
}}
\vspace{-5mm}
\label{tab:quantitative_pointcloud_reconstruction}
\end{center}
\end{table}

\subsection{Datasets and baselines}

\noindent\textbf{Datasets.} 
We train LISA on a non-released version of the InterHand2.6M dataset~\cite{interhand:moon:eccv20}, which contains multi-view sequences showing hands of 27 users. In total, there are 5804 multi-view frames and 131k images with the resolution of $1024\times667\,\text{px}$. Every frame has $\sim$22 views on average, two of which were not used for training and left for validation.
The dataset also provides a pseudo ground truth of the 3D joints, and we remove background in all images using hand masks obtained by a Mask R-CNN model~\cite{maskrcnn} provided by the authors of the dataset.
The geometry prior
is learned on the 3DH dataset~\cite{diform:xu:arxiv21} which contains sequences of 3D scans of 183 users (we use the same training/test split of 150/33 users proposed by the authors).
For evaluating hand reconstruction from point clouds, we use the test split of the MANO dataset~\cite{mano:romero:tog17}, which includes 50 3D scans of a single user, and the test set of 3DH, which includes scans of 33 users. For hand reconstruction from images, we use the DeepHandMesh dataset~\cite{dhm:moon:eccv20}, which is annotated with ground-truth 3D hand scans.

\customparagraph{Evaluated hand models.} 
As \Modelname~is the first neural model able to simultaneously represent hand geometry and texture, there are no published methods that would be directly comparable. To define baselines, we have therefore re-implemented several recent methods based on articulated implicit representations from the related field of human body modeling.
We adapt NASA~\cite{nasa:deng:2020} and NARF~\cite{narf:iccv21} to our setup by changing their geometry representation to signed distance fields, adding a positional encoding to NASA, and duplicating their geometry MLPs to predict also color. We train these methods on the InterHand2.6M dataset~\cite{interhand:moon:eccv20} with supervision on the skinning weights.
We did not manage to extend SNARF~\cite{snarf:chen:iccv21}, as it relies on an intermediate non-differentiable optimization during the forward pass that impedes calculating the output gradient with respect to the input points, which is necessary for applying the Eikonal loss.
We also compare to the original MANO model and to our implementation of VolSDF parameterized by the pose, shape and color vectors, but which does not consider a per-bone reasoning.
Besides, we ablate the following versions of the proposed model: the full model when trained with images and the geometric prior (LISA-full), a version trained solely with images (LISA-im), and a version trained only with the geometric prior (LISA-geom).

\subsection{Shape reconstruction from point clouds}
Table~\ref{tab:quantitative_pointcloud_reconstruction} summarizes the results of hand reconstruction from point clouds from the 3DH and MANO datasets. As the evaluation metrics, we report the vertex-to-vertex (V2V) and vertex-to-surface (V2S) distances (in millimeters). We compute these metrics in both directions, \ie from the reconstruction to the scan and the other way around. For a fair comparison, all reconstructions from all methods based on implicit representations are obtained with the same Marching Cubes resolution ($256\times256\times256$). Since MANO uses a mesh with only 778 vertices, we subdivide its reconstructed surface into $\sim$100k vertices.

The results show that LISA-im consistently outperforms the other methods when only images are used for training. Adding the geometric prior (LISA-full) yields a significant boost in performance. %
When the model is trained solely with the geometric prior (LISA-geom), it yields even lower errors than when trained using both the geometric prior and images (LISA-full). This is because we segmented out the hand in the training images %
and LISA-full and LISA-im therefore learned to close the surface right after the wrist. This spurious surface increases the measured error.

\input{figure_reconstruction_mano}

Figure~\ref{fig:qualitative_results_pointclouds} visualizes examples of the reconstructions.
Clear artifacts can be seen in most implicit models, except of LISA-full and the parametric MANO model.

\setlength{\tabcolsep}{0pt}
\begin{table}[t!]
\footnotesize
\begin{center}
\begin{tabularx}{\columnwidth}{l Y Y Y Y Y Y Y Y Y}
\toprule
& \multicolumn{3}{c}{1 view} & \multicolumn{3}{c}{2 views} & \multicolumn{3}{c}{4 views} \\

\cmidrule(lr){2-4} \cmidrule(lr){5-7} \cmidrule(lr){8-10}

Method  & V2V & V2S & PSNR  & V2V & V2S & PSNR  & V2V & V2S & PSNR \\

\cmidrule{1-10}

MANO~\cite{mano:romero:tog17}  & ~13.81 & ~8.93 & - & - & - & - & - & - & - \\
DHM~\cite{dhm:moon:eccv20} & ~9.86 & ~6.55 &  - & - &  - & - & - & - & - \\

VolSDF~\cite{volsdf:yariv:arxiv21} & 7.15 & 7.06 & 23.19 & 7.15 & 7.10 & 22.63 & 7.27 & 7.18 & 25.05  \\
NASA~\cite{nasa:deng:2020}  & 5.89 & 5.79 &  {\bf 25.20} & 5.11 & 4.99 & 25.17 & 5.04 & 4.91 & 25.18  \\
NARF~\cite{narf:iccv21}  & 7.44 & 7.35 & 24.11 &  7.45 & 7.36 & 28.48 & 7.93 & 7.85 & 29.89   \\
LISA-im  &  {\bf 5.48} & {\bf 5.36} & 25.04 & {\bf 3.86} & {\bf 3.72} & {\bf 29.84} & {\bf 3.62} & {\bf 3.47} & {\bf 30.21}  \\
\cmidrule{1-10}
LISA-full & 3.84 & 3.68 & 25.43 & 3.70 &  3.56 & 29.40 & 3.53 & 3.38 & 29.69  \\

\bottomrule
\end{tabularx}
\caption{\small{\textbf{Shape and color reconstruction of DHM~\cite{dhm:moon:eccv20} images.}
The 3D shape reconstructions are evaluated by the vertex-to-vertex and vertex-to-surface distances (in mm) and color renderings of the hand models in novel views are evaluated by the PSNR metric~\cite{nerf:eccv20}. Scores for MANO and DeepHandMesh (DHM) were taken from~\cite{dhm:moon:eccv20}.
We also report metrics for 1, 2 or 4 views, out of the 5 available images in~\cite{dhm:moon:eccv20}. In the same conditions, LISA-im outperforms all other methods trained on images only. When trained also with the geometry prior (LISA-full), it achieves an additional boost that is most noticeable in the single-view setup.
}}
\vspace{-1mm}
\label{tab:quantitative_scans}
\end{center}
\end{table}

\setlength{\tabcolsep}{0pt}
\begin{table}[t!]
\footnotesize
\begin{center}
\begin{tabularx}{\columnwidth}{l Y Y Y Y Y Y Y Y Y}
\toprule
& \multicolumn{3}{c}{1 view} & \multicolumn{3}{c}{2 views} & \multicolumn{3}{c}{4 views} \\
\cmidrule(lr){2-4} \cmidrule(lr){5-7} \cmidrule(lr){8-10}
&
{\scriptsize PSNR$\uparrow$} &
{\scriptsize SSIM$\uparrow$} &
{\scriptsize LPIPS$\downarrow$} &
{\scriptsize PSNR$\uparrow$} &
{\scriptsize SSIM$\uparrow$} &
{\scriptsize LPIPS$\downarrow$} &
{\scriptsize PSNR$\uparrow$} &
{\scriptsize SSIM$\uparrow$} &
{\scriptsize LPIPS$\downarrow$} \\
\cmidrule{1-10}
VolSDF
& 23.01 & 0.92 & 0.12 & 23.36 & 0.87 & 0.11 & 23.89 & 0.93 & 0.11  \\
NASA
& 26.90 & 0.95 & 0.07 & 28.16 & 0.96 & 0.06 & 28.44 & 0.96 & 0.06  \\
NARF
& 28.42 & 0.95 & 0.09 & 28.49 & 0.96 & 0.72 & 28.59 & 0.96 & 0.08 \\
LISA-im  & 27.45 & 0.95 & 0.08 & 28.40 & 0.95 & 0.07 & 28.27 & 0.95 & 0.07   \\
LISA-full  & 27.07 & 0.95 & 0.06 & 27.69 & 0.96 & 0.06 & 28.27 & 0.96 & 0.05 \\
\bottomrule
\end{tabularx}
\caption{
\small{\textbf{Color reconstruct. from InterHand2.6M images~\cite{interhand:moon:eccv20}.}
All
models achieve comparable performance
in terms of PSNR and SSIM (measuring the pixel error; higher is better) and LPIPS (measuring the overall perceptual similarity; lower is better)~\cite{nerf:eccv20}.
}
}
\vspace{-1mm}
\label{tab:quantitative_image_reconstruction}
\end{center}
\end{table}

\input{figure_reconstruction_interhand}
\input{figure_qualitative_freihand}
 
\subsection{Shape and color reconstruction from images}
Table~\ref{tab:quantitative_scans} evaluates the hand models on the task of 3D reconstruction from single and multiple views on the DeepHandMesh dataset~\cite{dhm:moon:eccv20}.
Among methods trained on images only, LISA-im is consistently superior in 3D shape reconstruction, and its performance is further boosted when the geometric prior is employed (LISA-full).

Color reconstruction from DeepHandMesh images~\cite{dhm:moon:eccv20} is evaluated in Table~\ref{tab:quantitative_scans} by the PSNR metric~\cite{nerf:eccv20} calculated on renderings of the hand model from novel views. LISA-im is slightly superior in this metric, with exception of the case when a single image is used for the reconstruction, where the performance of LISA-im is on par with NASA. Color reconstruction from InterHand2.6M images~\cite{interhand:moon:eccv20} is evaluated in Table~\ref{tab:quantitative_image_reconstruction} by the PSNR, SSIM and LPIPS~\cite{nerf:eccv20} metrics. In this case, all methods are fairly comparable in terms of rendering quality, which we suspect is likely
due to noisy hand masks used in training.

Qualitative results are shown in Figure~\ref{fig:qualitative_results_image}.
Additionally, we demonstrate in Figure~\ref{fig:freihand} that LISA can reconstruct hands from images in the wild, even in cases where the hand is partially occluded by an object. We refer the reader to the supplementary material for additional qualitative results.

\subsection{Inference speed}

To reconstruct the LISA hand model from one or multiple views, we first optimize the pose parameters for 1k iterations, after which we jointly optimize shape, pose and color parameters for additional 5k iterations. This process takes approximately 5 minutes. 
After converging, we reconstruct meshes at the resolution of $128^3$, which takes around 5 seconds, or render novel views in approximately one minute. These measurements were made on $1024\times667\,\text{px}$ images with a single Nvidia Tesla P100 GPU. The inference speed is similar for the NASA and NARF models, which also perform per-bone predictions. VolSDF is $\sim$2 faster due to the fact that it only uses a single MLP.

\vspace{-5pt}

\section{Conclusion}

We have introduced LISA, a novel neural representation of textured hands, which we learn by combining volume rendering approaches with a hand geometric prior. The resulting model is the first one to allow full and independent control of pose, shape and color domains. We show the utility of LISA in two challenging problems, hand reconstruction from point-clouds and hand reconstruction from images. In both of these applications we obtain highly accurate 3D shape reconstructions, achieving a sub-millimeter error in point-cloud fitting and surpassing the evaluated baselines by large margins. This level of accuracy is not possible to achieve with low-resolution parametric meshes such as MANO~\cite{mano:romero:tog17} or with models representing a single person such as DeepHandMesh~\cite{dhm:moon:eccv20}.
Future research directions include exploring temporal consistency for tracking applications, eliminating the need of rough 2D/3D pose of the hand skeleton and foreground mask at inference, improving the run-time efficiency, or enhancing the expressiveness in terms of high-frequency textural details while maintaining the generalization capability.

\noindent{\bf Acknowledgements}: 
This work is supported in part by the Spanish government with the project MoHuCo PID2020-120049RB-I00.

{\small
\bibliographystyle{ieee_fullname}
\bibliography{references.bib}
}

\end{document}